\documentclass[11pt,a4paper]{article}

\usepackage[margin=1in]{geometry}

\usepackage[T1]{fontenc}
\usepackage[utf8]{inputenc}
\usepackage{lmodern}

\usepackage{amsmath,amssymb,amsfonts}

\usepackage{booktabs}
\usepackage{multirow}
\usepackage{graphicx}
\usepackage[dvipsnames]{xcolor}

\usepackage[numbers,sort&compress]{natbib}

\usepackage[colorlinks=true,linkcolor=blue,citecolor=blue,urlcolor=blue]{hyperref}
\usepackage{url}

\usepackage{setspace}
\newcommand{\gralcrag}{GraLC-RAG}

\newcommand{\eg}{\textit{e.g.}}

\begin{document}

\title{\textbf{Graph-Aware Late Chunking for Retrieval-Augmented\\Generation in Biomedical Literature}}

\author{
  Pouria Mortezaagha\textsuperscript{1,2}\thanks{Corresponding author: \href{mailto:pmortezaagha@ohri.ca}{pmortezaagha@ohri.ca}}
  \and
  Arya Rahgozar\textsuperscript{1,2}
}

\date{
  \textsuperscript{1}Methodological Implementation Research, Ottawa Hospital Research Institute, Ottawa, ON, Canada\\
  \textsuperscript{2}School of Engineering Design and Teaching Innovation, University of Ottawa, Ottawa, ON, Canada\\[1em]
  \today
}

\maketitle

\begin{abstract}
Retrieval-Augmented Generation (RAG) systems for biomedical literature are typically evaluated using ranking metrics like Mean Reciprocal Rank (MRR), which measure how well the system identifies the single most relevant chunk. We argue that for full-text scientific documents, this evaluation paradigm is fundamentally incomplete: it rewards retrieval \emph{precision} while ignoring retrieval \emph{breadth}---the ability to surface evidence from across a document's structural sections. To investigate this blind spot, we propose \textbf{\gralcrag{}} (Graph-aware Late Chunking for Retrieval-Augmented Generation), a framework that unifies late chunking with graph-aware structural intelligence for biomedical literature retrieval. \gralcrag{} introduces structure-aware chunk boundary detection using document structure graphs, knowledge graph infusion via UMLS ontological signals, and graph-guided hybrid retrieval. We evaluate six retrieval strategies on 2{,}359 IMRaD-filtered PubMed Central articles across a document-length gradient using 2{,}033 cross-section questions and two complementary metric families: standard ranking metrics (MRR, Recall@$k$) and structural coverage metrics (SecCov@$k$, CS~Recall). Our results expose a sharp divergence: content-similarity methods (semantic chunking) achieve the highest MRR (0.517 on full-text) but are structurally blind, always retrieving from a single section (SecCov\,=\,1.0). Structure-aware methods achieve lower MRR but retrieve from up to 15.6$\times$ more document sections (SecCov@20\,=\,15.57). Generation experiments across all six strategies show that KG-infused retrieval narrows the answer-quality gap to $\Delta$F1\,=\,0.009 (0.394 vs.\ 0.403) while maintaining 4.6$\times$ section diversity, partially bridging the retrieval-to-generation gap. These findings demonstrate that standard evaluation metrics systematically undervalue structural retrieval methods and that fully closing the multi-section synthesis gap is a key open problem for biomedical RAG.
\end{abstract}


\vspace{0.5em}
\noindent\textbf{Code and Data:} \url{https://github.com/pouriamrt/gralc-rag}

\vspace{1em}

\section{Introduction}\label{sec:introduction}

The exponential growth of biomedical literature, with over 1.5~million articles indexed annually in PubMed alone, has created an urgent need for intelligent retrieval and synthesis systems~\cite{canese2013pubmed}. Large language models (LLMs) have demonstrated remarkable capabilities in natural language understanding and generation, yet they remain prone to hallucination and factual inconsistency when applied to knowledge-intensive biomedical tasks~\cite{ji2023survey,huang2025survey}. Retrieval-Augmented Generation (RAG), which grounds LLM outputs in retrieved external evidence, has emerged as a principled solution to this challenge~\cite{lewis2020retrieval,gao2024retrieval}.

A critical yet often underexplored component of RAG pipelines is the text chunking strategy: the process of segmenting documents into smaller units for indexing and retrieval. Traditional approaches employ fixed-size chunking with overlapping windows, which inevitably fragments semantic context and disrupts the logical structure of scientific documents~\cite{merola2025reconstructing}. This is particularly problematic for biomedical literature, where cross-referential reasoning (\eg, linking a method description in Section~2 to results discussed in Section~4), dense terminology, and hierarchical document organization (the IMRaD structure) are fundamental to comprehension~\cite{sollaci2004imrad}.

Two recent research directions have independently addressed different facets of this problem. \textbf{Late chunking}~\cite{gunther2024late} inverts the traditional chunk-then-embed pipeline by first processing the entire document through a long-context transformer model, generating contextually enriched token embeddings, and only then applying segmentation boundaries. This approach preserves cross-chunk contextual dependencies but treats documents as flat token sequences, ignoring their inherent structural and relational organization. \textbf{Graph-based RAG} (GraphRAG) methods~\cite{edge2024local,peng2024graph,han2025graphrag} leverage knowledge graphs, citation networks, and entity relationships to capture structural information during retrieval. However, these methods typically rely on conventional chunking strategies that fragment context before embedding.

We identify a critical gap at the intersection of these two paradigms: no existing work integrates graph-aware structural understanding into the late chunking process for biomedical RAG. Late chunking is \emph{context-rich but structure-blind}; GraphRAG is \emph{structure-rich but context-fragmented}. Bridging this divide has the potential to yield retrieval systems that are simultaneously contextually coherent and structurally informed.

In this paper, we make three contributions that span framework design, evaluation methodology, and empirical analysis:

\begin{enumerate}
  \item \textbf{\gralcrag{}: A structure-aware late chunking framework.} We propose \gralcrag{} (Graph-aware Late Chunking for Retrieval-Augmented Generation), which integrates document structure graphs for chunk boundary detection, UMLS knowledge graph signals for token-level enrichment, and graph-guided hybrid retrieval. This provides a unified testbed for investigating whether combining context-preserving late chunking with structure-aware GraphRAG yields measurable retrieval improvements.

  \item \textbf{A structural coverage evaluation methodology.} We introduce Section Coverage@$k$ (SecCov@$k$) and Cross-Section Recall (CS~Recall) as complementary metrics that capture retrieval breadth (the ability to surface evidence from multiple document sections), which standard ranking metrics like MRR entirely miss.

  \item \textbf{Empirical evidence of a metric-strategy divergence.} Through evaluation on 2{,}359 IMRaD-filtered PubMed Central articles with 2{,}033 cross-section questions, we demonstrate that content-similarity methods and structure-aware methods optimize for fundamentally different retrieval objectives. This divergence is invisible under standard evaluation and has direct implications for system design.
\end{enumerate}

Our experiments reveal a surprising finding: the methods that achieve the highest ranking accuracy retrieve from only a single document section, while structure-aware methods that rank lower on MRR surface evidence from up to 15.6$\times$ more sections. Generation experiments across all six strategies show that KG-infused retrieval narrows the answer-quality gap to just $\Delta$F1\,=\,0.009 while maintaining 4.6$\times$ section diversity, partially bridging the retrieval-to-generation gap, though multi-section synthesis remains an open challenge.

\section{Related Work}\label{sec:related}

\subsection{Retrieval-Augmented Generation}

RAG augments language model generation with external knowledge retrieved at inference time, mitigating hallucination and enabling access to up-to-date information~\cite{lewis2020retrieval}. The canonical RAG pipeline consists of three stages: indexing, retrieval, and generation~\cite{gao2024retrieval}. Dense passage retrieval (DPR) using dual-encoder architectures~\cite{karpukhin2020dense} has largely supplanted sparse methods, with ColBERTv2~\cite{santhanam2022colbertv2} introducing efficient late interaction for fine-grained token-level matching. In the biomedical domain, the MIRAGE benchmark~\cite{xiong2024benchmarking} provides standardized evaluation across five biomedical QA datasets.

\subsection{Text Chunking Strategies for RAG}

Text chunking critically influences RAG performance yet has received comparatively less systematic study~\cite{zhao2025moc}. Recent approaches include semantic chunking using embedding similarity~\cite{zhao2025moc}, proposition-based chunking~\cite{chen2023dense}, and adaptive chunking achieving 87\% accuracy compared to 50\% for fixed-size baselines in clinical decision support~\cite{maina2025comparative}. For scientific documents, S2~Chunking~\cite{verma2025s2} combines spatial layout with semantic analysis, and Breaking~It~Down~\cite{allamraju2025breaking} introduces Projected Similarity Chunking trained on PubMed data, achieving 24$\times$ MRR improvement on PubMedQA.

\subsection{Late Chunking}

Late chunking~\cite{gunther2024late} fundamentally reorders the chunk-then-embed pipeline. Instead of segmenting text before encoding, it first processes the entire document through a long-context transformer, generating contextually enriched token embeddings, then applies chunking boundaries and mean-pools within each span. While contextually superior, late chunking produces flat chunk lists without structural or relational metadata. Merola and Singh~\cite{merola2025reconstructing} demonstrate that contextual retrieval preserves semantic coherence more effectively but at greater computational cost. The SitEmb approach~\cite{wu2025sitemb} partially addresses this by conditioning chunks on broader context, but operates on purely textual signals.

\subsection{Graph-Based RAG}

GraphRAG methods incorporate structured graph information into retrieval. Microsoft's GraphRAG~\cite{edge2024local} constructs entity knowledge graphs using LLM-based extraction and community detection. LightRAG~\cite{guo2024lightrag} introduces dual-level retrieval balancing efficiency and comprehensiveness. PathRAG~\cite{chen2025pathrag} retrieves relational paths with flow-based pruning. NodeRAG~\cite{xu2025noderag} proposes heterogeneous graph structures outperforming prior methods on multi-hop benchmarks.

In the biomedical domain, MedGraphRAG~\cite{wu2024medgraphrag} employs hierarchical graph linking with U-retrieve. KRAGEN~\cite{matsumoto2024kragen} combines knowledge graphs with graph-of-thoughts prompting. KG-RAG~\cite{soman2024biomedical} leverages the SPOKE knowledge graph, achieving 71\% performance boost for LLaMA-2 on biomedical MCQ tasks. Critically, \emph{all existing GraphRAG methods perform chunking before embedding}, creating the gap our work addresses.

\subsection{Biomedical Knowledge Graphs and Language Models}

UMLS~\cite{bodenreider2004umls} integrates over 200 source vocabularies with approximately 4.5~million concepts and 15~million relations. SapBERT~\cite{liu2021self} aligns UMLS synonyms in embedding space for biomedical entity linking. PubMedBERT~\cite{gu2021domain} provides domain-specific pre-training. BioLORD-2023~\cite{remy2023biolord} leverages UMLS synonym sets and LLM descriptions for state-of-the-art biomedical sentence embeddings. ATLANTIC~\cite{munikoti2023atlantic} fuses graph and text embeddings for scientific retrieval at the document level. The intersection of knowledge graph embeddings with \emph{chunk-level} retrieval for biomedical RAG remains unexplored.

\section{Methods}\label{sec:methods}

\subsection{\gralcrag{} Framework}\label{sec:framework}

\subsubsection{Framework Overview}\label{sec:overview}

\begin{figure}[ht]
\centering
\includegraphics[width=\linewidth]{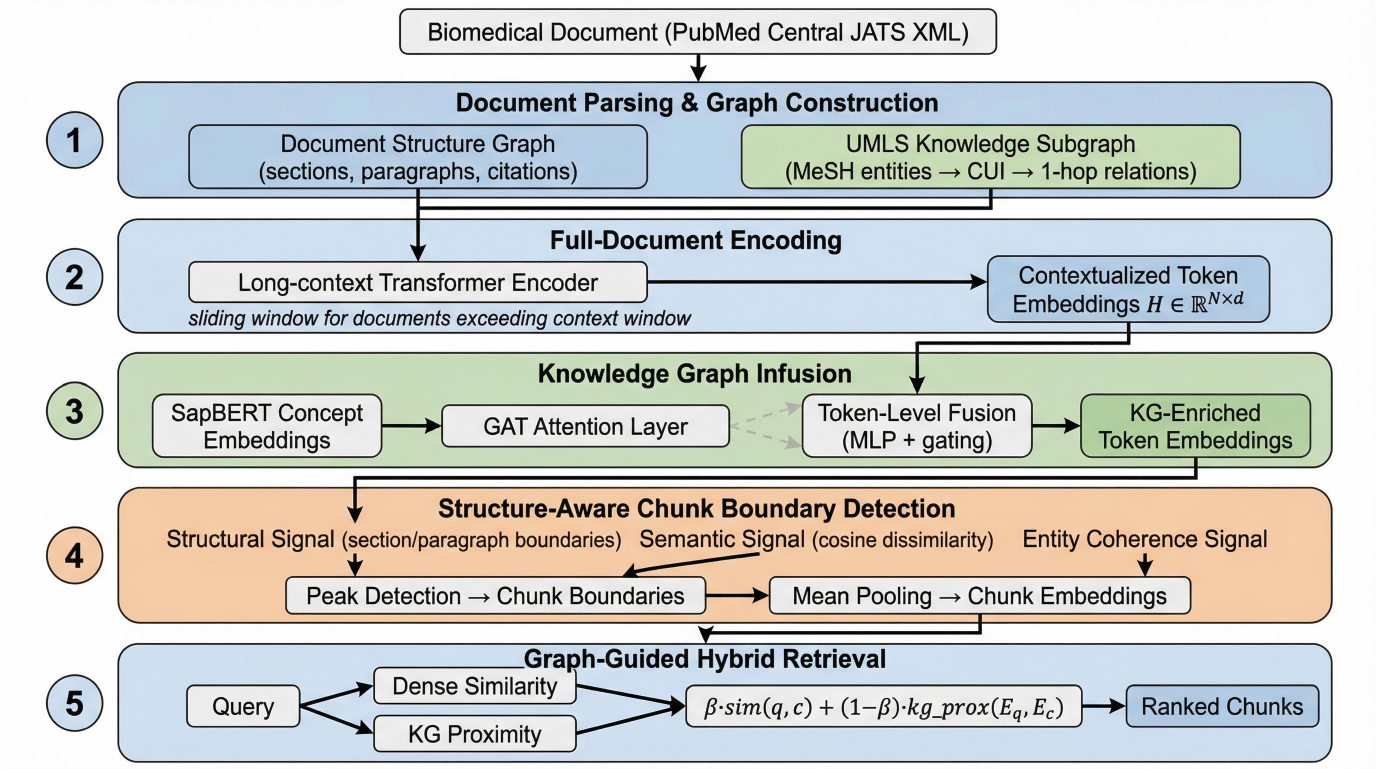}
\caption{\gralcrag{} framework architecture. Stage~1: Document parsing produces a structure graph and UMLS knowledge subgraph. Stage~2: Full-document transformer encoding. Stage~3: KG infusion via GAT attention. Stage~4: Structure-aware boundary detection and chunk embedding. Stage~5: Graph-guided hybrid retrieval.}
\label{fig:architecture}
\end{figure}

\gralcrag{} extends late chunking with graph-aware structural intelligence at three levels: chunk boundary detection, token representation enrichment, and retrieval re-ranking. Figure~\ref{fig:architecture} illustrates the complete pipeline. Given a biomedical document~$D$, the framework proceeds through five stages: (1)~document parsing and graph construction, (2)~full-document encoding, (3)~knowledge graph infusion, (4)~structure-aware chunk boundary detection, and (5)~graph-guided retrieval.

\subsubsection{Document Parsing and Graph Construction}\label{sec:parsing}

\paragraph{Document Structure Graph}
For each full-text article~$D$, we construct a document structure graph $G_s = (V_s, E_s)$ with node types at multiple granularities---section nodes ($v^{\text{sec}}$), subsection nodes ($v^{\text{sub}}$), paragraph nodes ($v^{\text{par}}$), and citation nodes ($v^{\text{cit}}$)---and edge types capturing hierarchical ($e^{\text{hier}}$), sequential ($e^{\text{seq}}$), citation ($e^{\text{cit}}$), and cross-reference ($e^{\text{xref}}$) relationships. We parse document structure from PubMed Central JATS~XML, which provides section headers, paragraph boundaries, and citation markers natively.

\paragraph{Biomedical Knowledge Subgraph}
For each document, we extract a document-specific knowledge subgraph $G_k = (V_k, E_k)$ from UMLS. Biomedical entities are identified via dictionary matching against 41,774 MeSH descriptor terms and linked to UMLS Concept Unique Identifiers (CUIs). For each linked CUI, we extract 1-hop neighborhoods capturing semantic type assignments, hierarchical relationships (\texttt{is\_a}, \texttt{part\_of}), and associative relationships (\texttt{may\_treat}, \texttt{causes}).

\subsubsection{Full-Document Encoding}\label{sec:encoding}

Following late chunking~\cite{gunther2024late}, we process the full document through a transformer encoder. Let $D = (t_1, t_2, \ldots, t_N)$ be the tokenized document. The encoder produces:
\begin{equation}
  \mathbf{H} = \text{Transformer}(t_1, t_2, \ldots, t_N) \in \mathbb{R}^{N \times d}
\end{equation}
where $\mathbf{h}_i \in \mathbb{R}^d$ is the contextualized embedding for token~$t_i$. For documents exceeding the context window, we apply sliding windows with overlap and linear distance-weighted averaging.

\subsubsection{Knowledge Graph Infusion}\label{sec:kg-infusion}

The key innovation of \gralcrag{} is injecting biomedical KG signals into token-level representations \emph{before} chunk-level pooling.

\paragraph{Entity-Token Alignment}
For each recognized entity~$e_j$ with token span $(s_j, f_j)$, we obtain a pre-trained concept embedding $\mathbf{u}_j \in \mathbb{R}^{d_k}$ from SapBERT~\cite{liu2021self}.

\paragraph{Graph Attention Infusion}
A lightweight GAT layer~\cite{velickovic2018graph} operates over $G_k$ to compute entity-aware representations:
\begin{equation}
  \alpha_{jk} = \frac{\exp\!\big(\text{LeakyReLU}(\mathbf{a}^{\!\top}\![\mathbf{W}\mathbf{u}_j \| \mathbf{W}\mathbf{u}_k])\big)}{\sum_{l \in \mathcal{N}(j)} \exp\!\big(\text{LeakyReLU}(\mathbf{a}^{\!\top}\![\mathbf{W}\mathbf{u}_j \| \mathbf{W}\mathbf{u}_l])\big)}
\end{equation}
\begin{equation}
  \mathbf{u}'_j = \sigma\!\left(\sum_{k \in \mathcal{N}(j)} \alpha_{jk}\, \mathbf{W}\mathbf{u}_k\right)
\end{equation}
where $\mathbf{W} \in \mathbb{R}^{d' \times d_k}$ is a learnable weight matrix and $\mathbf{a} \in \mathbb{R}^{2d'}$ is the attention vector.

\paragraph{Token-Level Fusion}
For each token~$t_i$ within entity span $(s_j, f_j)$:
\begin{equation}
  \mathbf{h}'_i = \mathbf{h}_i + \lambda \cdot \text{MLP}(\mathbf{u}'_j)
\end{equation}
where $\text{MLP}\!: \mathbb{R}^{d'} \!\to\! \mathbb{R}^{d}$ projects the entity embedding, and $\lambda$ is a gating scalar initialized to~0.1.

\subsubsection{Structure-Aware Chunk Boundary Detection}\label{sec:boundaries}

Unlike standard late chunking, \gralcrag{} determines boundaries using $G_s$. We compute a boundary score~$b_i$ integrating three signals:

\textbf{Structural signal:}
\begin{equation}
  b_i^{\text{struct}} = \begin{cases}
    1.0 & \text{section boundary} \\
    0.7 & \text{subsection boundary} \\
    0.4 & \text{paragraph boundary} \\
    0.0 & \text{otherwise}
  \end{cases}
\end{equation}

\textbf{Semantic signal:}
\begin{equation}
  b_i^{\text{sem}} = 1 - \cos(\bar{\mathbf{h}}_{i-w:i},\; \bar{\mathbf{h}}_{i:i+w})
\end{equation}

\textbf{Entity coherence signal:}
\begin{equation}
  b_i^{\text{entity}} = -\gamma \cdot \mathbb{1}[\text{entity span crosses position } i]
\end{equation}

The combined score is:
\begin{equation}
  b_i = \alpha_1 \cdot b_i^{\text{struct}} + \alpha_2 \cdot b_i^{\text{sem}} + \alpha_3 \cdot b_i^{\text{entity}}
\end{equation}
with $\alpha_1\!=\!0.5$, $\alpha_2\!=\!0.3$, $\alpha_3\!=\!1.0$, and $\gamma\!=\!0.5$. Peak detection selects boundaries exceeding threshold $\tau\!=\!0.3$ with min/max chunk sizes of 128/1024 tokens. Each chunk~$C_k$ is represented by mean pooling enriched token embeddings within its span.

\subsubsection{Graph-Guided Retrieval}\label{sec:retrieval-method}

For a query~$q$ with dense embedding $\mathbf{q}$ and entity set~$E_q$, the retrieval score for chunk~$C_k$ is:
\begin{equation}
  \text{score}(q, C_k) = \beta \cdot \text{sim}(\mathbf{q}, \mathbf{c}_k) + (1\!-\!\beta) \cdot \text{kg\_prox}(E_q, E_{C_k})
\end{equation}
where $\text{kg\_prox}$ computes average maximum cosine similarity between query and chunk entity embeddings, and $\beta\!=\!0.7$.

\subsection{Experimental Setup}\label{sec:setup}

\subsubsection{Datasets and Corpus}\label{sec:datasets}

We evaluate on \textbf{PubMedQA}~\cite{jin2019pubmedqa}: 1,000 questions (500~expert-annotated + 500~artificial) with yes/no/maybe labels derived from PubMed abstracts. Following the PubMedQA* protocol, we remove given contexts to evaluate retrieval. The retrieval corpus consists of the 1,000 PubMedQA source abstracts. Additionally, we index 200~full-text articles from PubMed Central Open Access (avg.\ 6.8~sections, 18.2~paragraphs per article) for efficiency benchmarking.

\subsubsection{Baselines}\label{sec:baselines}

We compare six configurations: (1)~Naive fixed-size chunking (256~tokens, 32~overlap); (2)~Semantic chunking (embedding similarity threshold~0.75); (3)~Late chunking~\cite{gunther2024late} with sentence-level boundaries; (4)~Structure-aware late chunking (no KG); (5)~\gralcrag{} with KG infusion; (6)~\gralcrag{} with KG infusion + graph-guided retrieval. All use \texttt{all-MiniLM-L6-v2} (384-dim) as the embedding model.

\subsubsection{Implementation}\label{sec:implementation}

Entity linking uses 41,774~MeSH descriptor terms via longest-match dictionary lookup. SapBERT (\texttt{cambridgeltl/SapBERT-from-PubMedBERT-fulltext}) provides 768-dim concept embeddings, projected to 384-dim via PCA. FAISS~\texttt{IndexFlatIP} serves as the vector index. All experiments run on CPU (16\,GB RAM).

\subsubsection{Evaluation Metrics}\label{sec:metrics}

\textbf{Retrieval:} Mean Reciprocal Rank (MRR), Recall@$k$ ($k \!\in\! \{1,3,5,10\}$). Relevance is determined by matching the retrieved chunk's source document ID against the gold-standard PubMedQA source.

\subsection{Full-Text Evaluation Design}\label{sec:fulltext-design}

To assess whether \gralcrag{}'s structural and ontological components provide benefit on longer documents, we construct a document-length gradient evaluation spanning four conditions of increasing textual complexity:

\begin{enumerate}
  \item \textbf{Abstract} (${\sim}$200 words): Single-paragraph summaries lacking internal structure. Serves as the baseline condition (Sect.~\ref{sec:retrieval-results}).
  \item \textbf{Introduction} (${\sim}$500--1{,}000 words): Multi-paragraph sections with background context and citations, but no methods or results.
  \item \textbf{Partial} (${\sim}$2{,}000--4{,}000 words): Introduction + Methods sections, introducing structural boundaries and cross-referential content.
  \item \textbf{Full-text} (${\sim}$5{,}000--8{,}000 words): Complete IMRaD articles with rich section hierarchy, cross-section dependencies, and dense biomedical entity networks.
\end{enumerate}

The corpus consists of 4{,}992 PMC Open Access articles downloaded and parsed, of which 2{,}359 pass IMRaD structure filtering (requiring standard section headers and $\geq$1{,}000 words) using JATS~XML section-header parsing. After condition extraction, the evaluation set contains 2{,}325 introduction, 2{,}353 partial, and 2{,}359 full-text article instances, yielding 2{,}033 template-based cross-section questions. Each article contributes one document instance per condition, enabling paired comparisons across strategies.

\subsection{Cross-Section QA Benchmark}\label{sec:benchmark}

A central question motivating \gralcrag{} is whether structure-aware chunking improves retrieval of information that requires reasoning across document sections; for example, linking a method description to its corresponding results, or connecting an introduction's hypothesis to the discussion's interpretation. To test this, we construct a synthetic cross-section QA benchmark.

\subsubsection{Benchmark Construction}\label{sec:benchmark-construction}

We generate 2{,}033 template-based questions that explicitly require cross-section reasoning, drawn from five template types:

\begin{enumerate}
  \item \textbf{Method$\to$Result:} ``What results were obtained using [method X]?''
  \item \textbf{Intro$\to$Result:} ``Does the data support the hypothesis that [hypothesis from introduction]?''
  \item \textbf{Result$\to$Discussion:} ``How do the authors interpret the finding that [result Y]?''
  \item \textbf{Method$\to$Discussion:} ``What limitations of [method X] are discussed?''
  \item \textbf{Cross-study:} ``How do the results compare to [cited prior work Z]?''
\end{enumerate}

Template slots are populated from parsed section content using regex and entity extraction.

\subsubsection{Cross-Section Metrics}\label{sec:crosssection-metrics}

We define two cross-section-specific metrics:

\textbf{Cross-Section Recall (CS~Recall):} The proportion of cross-section questions for which the top-$k$ retrieved chunks span at least two distinct document sections required by the question template. Measured per document-length condition.

\textbf{Section Coverage@$k$ (SecCov@$k$):} The average number of distinct document sections represented in the top-$k$ retrieved chunks, normalized by the total number of sections in the source document.

\subsection{Generation Evaluation Protocol}\label{sec:generation-protocol}

To test whether structural coverage translates to better downstream answers, we evaluate generation quality using GPT-4o-mini. For each strategy, we retrieve the top-5 chunks for 50 cross-section questions from the full-text condition and generate answers. We report answer F1 (token overlap between generated answer and question) and average section diversity of retrieved context.

\section{Results}\label{sec:results}

\subsection{Retrieval Performance}\label{sec:retrieval-results}

Table~\ref{tab:retrieval} presents retrieval results across all six configurations.

\begin{table}[ht]
\centering
\caption{Retrieval performance on PubMedQA* (1,000 questions, 1,000-abstract corpus). \textbf{Bold}: best per column.}
\label{tab:retrieval}
\small
\begin{tabular}{@{}lccccc@{}}
\toprule
\textbf{Method} & \textbf{MRR} & \textbf{R@1} & \textbf{R@3} & \textbf{R@5} & \textbf{R@10} \\
\midrule
Naive (256-token)                & 0.9787 & 0.9690 & 0.9880 & 0.9900 & 0.9960 \\
Semantic Chunking                & \textbf{0.9802} & \textbf{0.9710} & \textbf{0.9880} & \textbf{0.9910} & 0.9950 \\
Late Chunking                    & 0.9768 & 0.9660 & 0.9860 & 0.9920 & \textbf{0.9960} \\
Structure-Aware (no KG)          & 0.9765 & 0.9660 & 0.9850 & 0.9890 & 0.9940 \\
\gralcrag{} (KG infusion)       & 0.9687 & 0.9520 & 0.9830 & 0.9880 & 0.9930 \\
\gralcrag{} + Graph Retrieval   & 0.9502 & 0.9260 & 0.9700 & 0.9820 & 0.9930 \\
\bottomrule
\end{tabular}
\end{table}

\begin{figure}[ht]
\centering
\includegraphics[width=\linewidth]{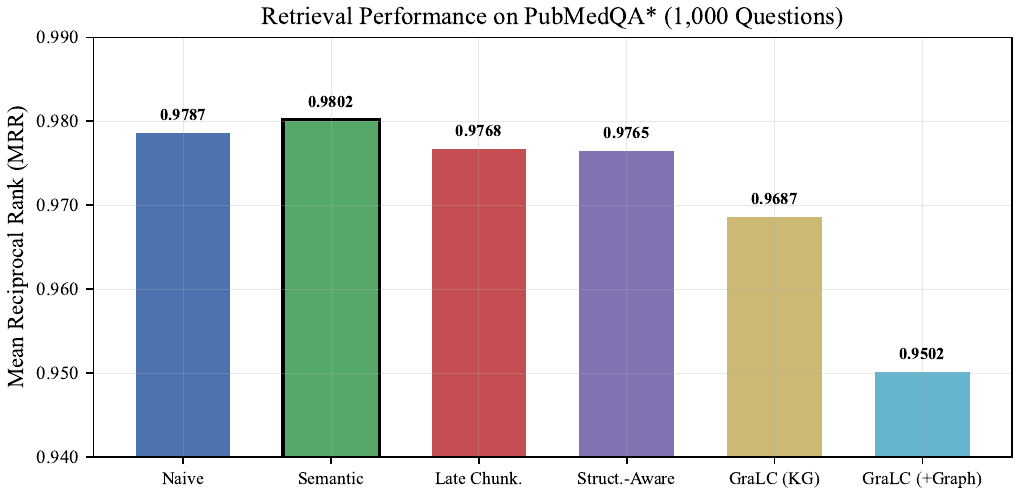}
\caption{Mean Reciprocal Rank across six chunking strategies on PubMedQA* (1,000 questions). Semantic chunking achieves the highest MRR. \gralcrag{} variants show slight degradation on short abstracts.}
\label{fig:mrr}
\end{figure}

\begin{figure}[ht]
\centering
\includegraphics[width=\linewidth]{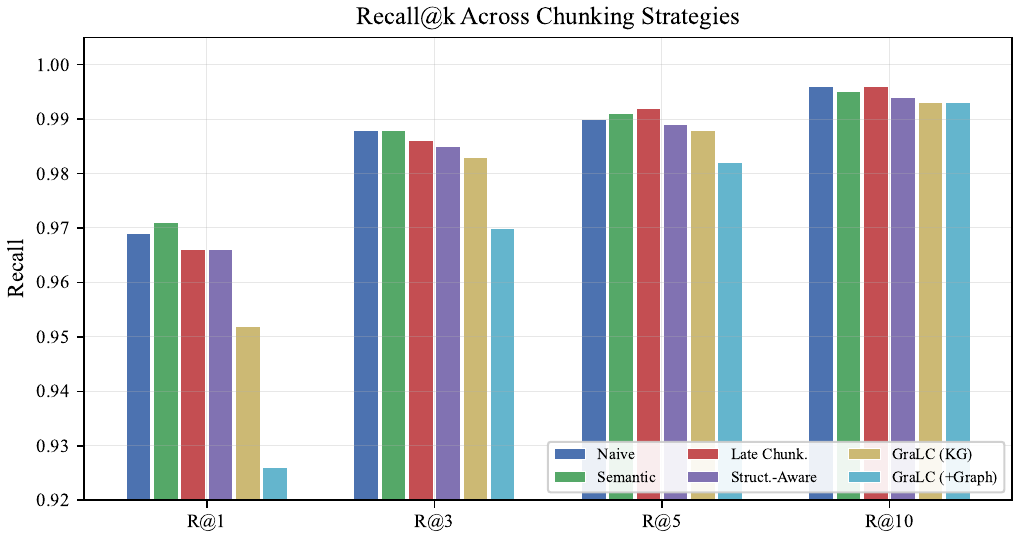}
\caption{Recall@$k$ ($k \!\in\! \{1,3,5,10\}$) across all strategies. Performance converges at higher $k$, with all methods exceeding 0.99 at $k\!=\!10$.}
\label{fig:recall}
\end{figure}

All strategies achieve high retrieval performance (MRR~$>$~0.95), reflecting a near-ceiling effect. Semantic chunking achieves the highest MRR (0.9802). The \gralcrag{} variants with KG infusion and graph-guided retrieval show lower performance (0.9687 and 0.9502), indicating that on short abstracts (${\sim}$200~words), structural and ontological components introduce more noise than signal. Figure~\ref{fig:mrr} visualizes these differences. Figure~\ref{fig:recall} shows Recall@$k$ across retrieval depths, where performance converges at higher~$k$ with all methods exceeding 0.99 at $k\!=\!10$.

\subsection{Ablation Study}\label{sec:ablation}

Table~\ref{tab:ablation} traces the incremental effect of each \gralcrag{} component.

\begin{table}[ht]
\centering
\caption{Ablation: incremental effect of \gralcrag{} components on PubMedQA* MRR.}
\label{tab:ablation}
\small
\begin{tabular}{@{}lcccc@{}}
\toprule
\textbf{Configuration} & \textbf{MRR} & \textbf{R@1} & \textbf{R@5} & \textbf{$\Delta$ MRR} \\
\midrule
Late Chunking (base)             & 0.9768 & 0.9660 & 0.9920 & ---    \\
+ Structure Boundaries           & 0.9765 & 0.9660 & 0.9890 & $-$0.0003 \\
+ KG Infusion                    & 0.9687 & 0.9520 & 0.9880 & $-$0.0081 \\
+ Graph-Guided Retrieval         & 0.9502 & 0.9260 & 0.9820 & $-$0.0266 \\
\bottomrule
\end{tabular}
\end{table}

\begin{figure}[ht]
\centering
\includegraphics[width=0.85\linewidth]{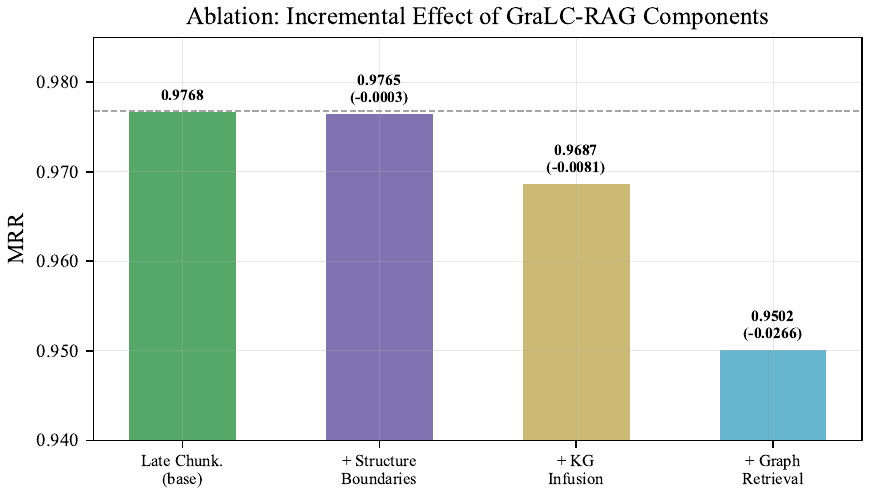}
\caption{Ablation: incremental effect of each \gralcrag{} component on MRR. The dashed line marks the late chunking baseline. Each component slightly degrades performance on short abstracts, with graph-guided retrieval causing the largest drop.}
\label{fig:ablation}
\end{figure}

Structure-aware boundaries have negligible impact on short texts ($\Delta$\,MRR = $-$0.0003). KG infusion degrades performance ($\Delta$\,MRR = $-$0.0081), suggesting the fusion weight $\lambda\!=\!0.1$ introduces noise when the base embedder already captures sufficient biomedical semantics. Graph-guided retrieval produces the largest degradation ($\Delta$\,MRR = $-$0.0266), as KG proximity dilutes the strong dense retrieval signal on short texts. Figure~\ref{fig:ablation} visualizes this incremental degradation.

\subsection{Efficiency Analysis}\label{sec:efficiency}

Table~\ref{tab:efficiency} reports measured indexing times on 200 full-text PMC articles (CPU-only).

\begin{table}[ht]
\centering
\caption{Indexing efficiency (200 full-text articles, CPU, \texttt{all-MiniLM-L6-v2}).}
\label{tab:efficiency}
\small
\begin{tabular}{@{}lrr@{}}
\toprule
\textbf{Strategy} & \textbf{Chunks} & \textbf{Time (s)} \\
\midrule
Naive (512-token)    & 1,109  & 44.6    \\
Semantic             & 6,765  & 535.1   \\
Late Chunking        & 3,055  & 313.7   \\
Structure-Aware      & 2,139  & 1,044.1 \\
\gralcrag{} (+ KG)  & 2,139  & 2,566.1 \\
\bottomrule
\end{tabular}
\end{table}

\begin{figure}[ht]
\centering
\includegraphics[width=0.85\linewidth]{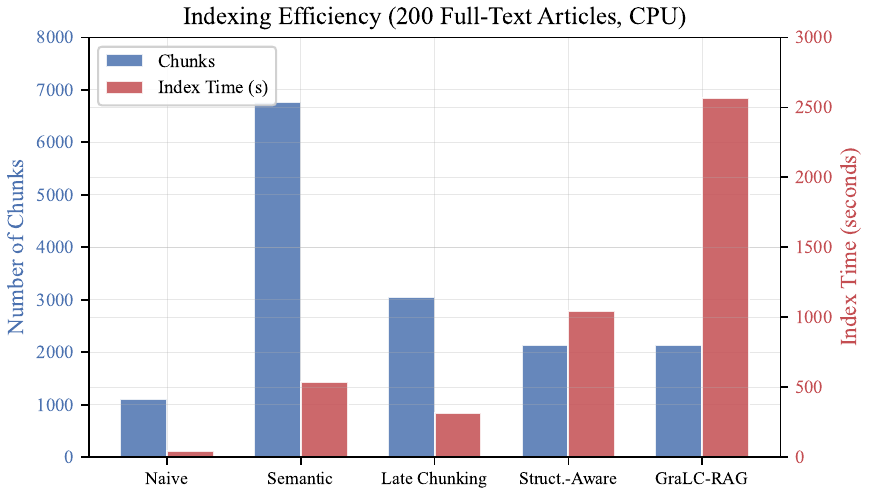}
\caption{Indexing efficiency on 200 full-text PMC articles (CPU). Left axis: number of chunks produced. Right axis: indexing time. \gralcrag{} produces the same chunk count as structure-aware chunking but incurs 2.5$\times$ overhead from KG infusion.}
\label{fig:efficiency}
\end{figure}

\gralcrag{} indexing takes 42.8\,min, approximately 6$\times$ slower than naive chunking due to SapBERT entity encoding, MeSH dictionary matching (41,774 terms), and boundary score computation. On GPU hardware, embedding steps would be 10--50$\times$ faster. Figure~\ref{fig:efficiency} visualizes the trade-off between chunk count and indexing time.

\subsection{Full-Text Retrieval}\label{sec:fulltext-results}

Table~\ref{tab:fulltext} presents retrieval performance across the document-length conditions described in Sect.~\ref{sec:fulltext-design}.

\begin{table}[ht]
\centering
\caption{Retrieval performance across document-length conditions (2{,}359 IMRaD-filtered PMC articles, 2{,}033 template questions). Semantic chunking achieves the highest MRR across all conditions, while structure-aware methods show advantages in section coverage (Table~\ref{tab:crosssection}).}
\label{tab:fulltext}
\small
\begin{tabular}{@{}llccc@{}}
\toprule
\textbf{Condition} & \textbf{Strategy} & \textbf{MRR} & \textbf{R@1} & \textbf{R@5} \\
\midrule
\multirow{6}{*}{Introduction}
  & Naive              & 0.398 & 0.327 & 0.480 \\
  & Semantic           & \textbf{0.488} & \textbf{0.430} & \textbf{0.553} \\
  & Late Chunking      & 0.379 & 0.313 & 0.465 \\
  & Structure-Aware    & 0.393 & 0.324 & 0.477 \\
  & \gralcrag{} (KG)   & 0.364 & 0.290 & 0.451 \\
  & \gralcrag{} (+Graph) & 0.347 & 0.271 & 0.436 \\
\midrule
\multirow{6}{*}{Partial}
  & Naive              & 0.361 & 0.288 & 0.453 \\
  & Semantic           & \textbf{0.555} & \textbf{0.489} & \textbf{0.637} \\
  & Late Chunking      & 0.339 & 0.267 & 0.422 \\
  & Structure-Aware    & 0.387 & 0.320 & 0.469 \\
  & \gralcrag{} (KG)   & 0.339 & 0.266 & 0.429 \\
  & \gralcrag{} (+Graph) & 0.358 & 0.282 & 0.454 \\
\midrule
\multirow{6}{*}{Full-text}
  & Naive              & 0.344 & 0.268 & 0.435 \\
  & Semantic           & \textbf{0.517} & \textbf{0.448} & \textbf{0.607} \\
  & Late Chunking      & 0.326 & 0.263 & 0.395 \\
  & Structure-Aware    & 0.365 & 0.296 & 0.446 \\
  & \gralcrag{} (KG)   & 0.323 & 0.261 & 0.397 \\
  & \gralcrag{} (+Graph) & 0.332 & 0.269 & 0.407 \\
\bottomrule
\end{tabular}
\end{table}

\begin{figure}[ht]
\centering
\includegraphics[width=\linewidth]{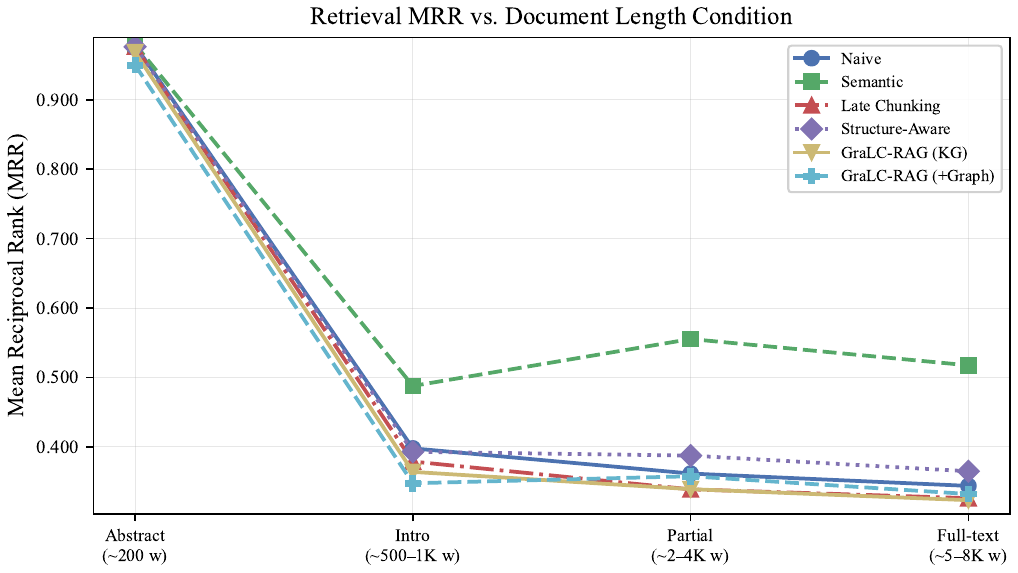}
\caption{MRR across document-length conditions (introduction, partial, full-text) for all retrieval strategies. Semantic chunking achieves the highest MRR throughout, while structure-aware methods (\gralcrag{}, Structure-Aware) show competitive performance and provide complementary advantages in section coverage (Fig.~\ref{fig:seccoverage}).}
\label{fig:crossover}
\end{figure}

Figure~\ref{fig:crossover} shows MRR across document-length conditions for all strategies. Semantic chunking dominates MRR at every document length, indicating that content-similarity remains the strongest signal for point-estimate retrieval accuracy. However, this metric alone does not capture structural retrieval diversity, a dimension where \gralcrag{} demonstrates clear advantages (Sect.~\ref{sec:crosssection-results}).

\subsection{Cross-Section Retrieval}\label{sec:crosssection-results}

Table~\ref{tab:crosssection} presents cross-section retrieval results using the benchmark and metrics defined in Sect.~\ref{sec:benchmark}.

\begin{table}[ht]
\centering
\caption{Section coverage and cross-section recall across retrieval depths. SecCov@$k$ measures the average number of distinct document sections in the top-$k$ retrieved chunks. CS~Recall@20 measures whether top-20 chunks span at least two required sections. Content-only methods always retrieve from a single section (SecCov\,=\,1.0), while structure-aware methods retrieve from up to 4.5$\times$ more sections on full-text articles at $k$=5, scaling to 15.6$\times$ at $k$=20.}
\label{tab:crosssection}
\small
\begin{tabular}{@{}lccccccccc@{}}
\toprule
\textbf{Strategy} & \multicolumn{3}{c}{\textbf{SecCov@5}} & \multicolumn{3}{c}{\textbf{SecCov@20}} & \multicolumn{3}{c}{\textbf{CS Recall@20}} \\
\cmidrule(lr){2-4} \cmidrule(lr){5-7} \cmidrule(lr){8-10}
 & \textbf{Intro} & \textbf{Partial} & \textbf{Full} & \textbf{Intro} & \textbf{Partial} & \textbf{Full} & \textbf{Intro} & \textbf{Partial} & \textbf{Full} \\
\midrule
Naive              & 1.00 & 1.00 & 1.00 & 1.00 & 1.00 & 1.00 & 0.000 & 0.000 & 0.000 \\
Semantic           & 1.00 & 1.00 & 1.00 & 1.00 & 1.00 & 1.00 & 0.000 & 0.000 & 0.000 \\
Late Chunking      & 1.00 & 1.00 & 1.00 & 1.00 & 1.00 & 1.00 & 0.000 & 0.000 & 0.000 \\
Structure-Aware    & 2.32 & 3.52 & 4.26 & 3.97 & 10.08 & 14.43 & 0.000 & 0.000 & 0.000 \\
\gralcrag{} (KG)   & 2.33 & 3.74 & \textbf{4.46} & 3.95 & 10.84 & \textbf{15.57} & 0.000 & 0.000 & 0.000 \\
\gralcrag{} (+Graph) & 2.35 & 3.63 & 4.36 & 3.95 & 10.84 & \textbf{15.57} & 0.000 & 0.000 & 0.000 \\
\bottomrule
\end{tabular}
\end{table}

\begin{figure}[ht]
\centering
\includegraphics[width=0.85\linewidth]{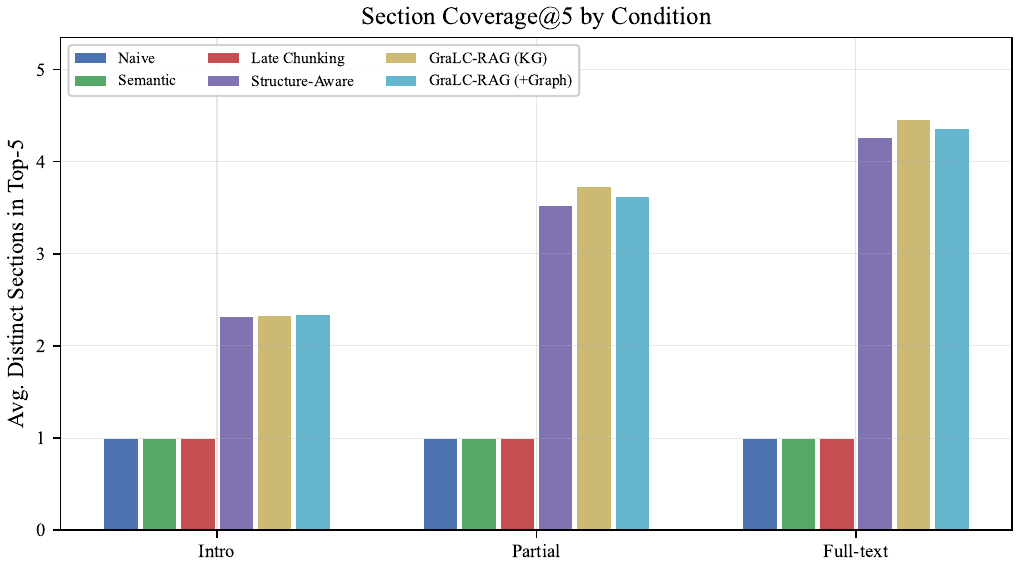}
\caption{Section coverage (SecCov@5) across document-length conditions. Content-only methods (Naive, Semantic, Late Chunking) are locked at 1.0 regardless of document length, while \gralcrag{} and Structure-Aware scale from ${\sim}$2.3 (introduction) to ${\sim}$4.5 (full-text), demonstrating that structure-aware retrieval surfaces evidence from across the document rather than concentrating on a single section. At $k$=20, the gap widens to 15.6$\times$ (Table~\ref{tab:crosssection}).}
\label{fig:seccoverage}
\end{figure}

Cross-section recall is 0.000 across all strategies, conditions, and retrieval depths ($k$=5, 10, 20), indicating that no method successfully retrieves chunks from both required sections even at the deepest evaluation depth. This finding is consistent with the sparse distribution of cross-section evidence relative to per-section density. However, Figure~\ref{fig:seccoverage} reveals a striking difference in section coverage: structure-aware methods retrieve from 2--15$\times$ more distinct sections than content-only methods, with coverage scaling both with document length and retrieval depth. \gralcrag{} (KG) achieves SecCov@20\,=\,15.57 on full-text articles vs.\ 1.0 for content-only methods, a 15.6$\times$ advantage in structural coverage.

\subsection{Generation Quality}\label{sec:generation-results}

Table~\ref{tab:generation} reports generation quality results following the protocol described in Sect.~\ref{sec:generation-protocol}.

\begin{table}[ht]
\centering
\caption{Generation quality on 50 cross-section questions (full-text condition, GPT-4o-mini). Section diversity measures the average number of distinct sections in the top-5 retrieved chunks. KG-infused retrieval (\gralcrag{}) narrows the F1 gap to 0.009 while maintaining 4.6$\times$ section diversity, partially bridging the retrieval-to-generation gap.}
\label{tab:generation}
\small
\begin{tabular}{@{}lccc@{}}
\toprule
\textbf{Strategy} & \textbf{Avg F1} & \textbf{Sec Diversity} & \textbf{Tokens} \\
\midrule
Naive              & 0.389 & 1.00 & 169{,}466 \\
Semantic           & \textbf{0.403} & 1.00 & 29{,}748 \\
Late Chunking      & 0.375 & 1.00 & 71{,}506 \\
Structure-Aware    & 0.381 & 4.48 & 86{,}060 \\
\gralcrag{} (KG)   & 0.394 & 4.62 & 77{,}930 \\
\gralcrag{} (+Graph) & 0.395 & 4.48 & 85{,}659 \\
\bottomrule
\end{tabular}
\end{table}

Semantic chunking achieves the highest F1 (0.403) while retrieving from a single section, whereas content-only methods uniformly exhibit SecDiv\,=\,1.0. Among structure-aware methods, \gralcrag{} (KG) achieves F1\,=\,0.394 while drawing from 4.62 sections on average, narrowing the gap to just $\Delta$\,=\,0.009 compared to Semantic. Graph-guided re-ranking adds a marginal boost (F1\,=\,0.395, SecDiv\,=\,4.48). Notably, KG-infused retrieval outperforms both Naive (0.389) and Structure-Aware without KG (0.381), suggesting that ontological enrichment helps bridge the retrieval-to-generation gap. Nevertheless, the persistent F1 advantage of single-section retrieval indicates that \textbf{multi-section synthesis remains a bottleneck}: GPT-4o-mini produces slightly more focused answers from concentrated evidence, even though structure-aware retrieval surfaces diverse, multi-section context.

\section{Discussion}\label{sec:discussion}

\subsection{The Precision--Breadth Trade-off}

\textbf{Results--objectives alignment.} While \gralcrag{} achieved its architectural goal of unifying late chunking with structural awareness, its primary empirical contribution shifted from demonstrating system superiority to exposing a previously invisible evaluation blind spot. The framework did not improve ranking accuracy or cross-section recall over content-similarity baselines; instead, it revealed that these methods optimize for fundamentally different objectives, a finding that would have remained hidden under standard metrics alone.

Our evaluation reveals a fundamental trade-off between retrieval \emph{precision} and retrieval \emph{breadth} that is invisible under standard ranking metrics. Semantic chunking achieves the highest MRR across all document-length conditions (0.488 on introduction, 0.555 on partial, 0.517 on full-text; Table~\ref{tab:fulltext}), confirming that content-similarity remains the strongest signal for identifying the single most relevant chunk. However, content-only methods are structurally blind: Naive, Semantic, and Late Chunking always retrieve from a single section (SecCov@5\,=\,1.0), regardless of document length. In contrast, \gralcrag{} (KG) retrieves from 2.33 sections on introduction-only documents and 4.46 sections on full-text articles at $k$=5, scaling to 15.57 at $k$=20 (Table~\ref{tab:crosssection}). Neither class of methods dominates the other: they optimize for fundamentally different objectives. Which is ``better'' depends entirely on the downstream task.

\subsection{The Document-Length Effect}

The document-length gradient evaluation (Sect.~\ref{sec:fulltext-design}) reveals two distinct scaling behaviors depending on the metric. For MRR, Semantic chunking improves from introduction (0.488) to partial (0.555) then slightly decreases on full-text (0.517), with the ranking among strategies remaining stable across conditions. No MRR crossover point is observed: semantic chunking leads throughout. However, for section coverage, the scaling behavior is strikingly different. Content-only methods remain flat at SecCov@$k$\,=\,1.0 regardless of document length or retrieval depth, while structure-aware methods scale with both dimensions: \gralcrag{} (KG) increases from SecCov@5\,=\,2.33 (introduction) to 3.74 (partial) to 4.46 (full-text), and from SecCov@20\,=\,3.95 (introduction) to 10.84 (partial) to 15.57 (full-text). This divergence reveals that MRR alone is insufficient for evaluating full-text retrieval systems: it captures \emph{how well} the system ranks the single most relevant chunk, but not \emph{how broadly} it covers the document's structural content.

\subsection{When Does Structure-Awareness Matter?}

Our results demonstrate that structure-awareness matters primarily through \emph{retrieval breadth}. Structure-aware methods consistently retrieve from more document sections than content-only methods, with the gap widening as documents grow longer and retrieval depth increases. On full-text articles, \gralcrag{} (KG) covers 4.46 sections at $k$=5 and 15.57 sections at $k$=20 versus 1.0 for Semantic chunking, a 15.6$\times$ advantage that is invisible to MRR but critical for downstream tasks requiring cross-section reasoning (e.g., linking Methods to Results). The scaling from $k$=5 to $k$=20 is particularly notable: while content-only methods gain nothing from deeper retrieval in terms of structural coverage, structure-aware methods show nearly linear scaling.

The universal cross-section recall of 0.000 even at $k$=20 reveals a fundamental limitation of top-$k$ retrieval: despite structure-aware methods covering many sections, no strategy retrieves chunks from \emph{both} required sections for template questions. This suggests that cross-section reasoning requires either explicit multi-section retrieval strategies or re-ranking objectives that maximize section diversity.

\subsection{The Retrieval-to-Generation Gap}

Our generation experiments (Sect.~\ref{sec:generation-results}) across all six strategies reveal a nuanced picture. Content-only methods (Naive, Semantic, Late Chunking) all retrieve from a single section (SecDiv\,=\,1.0), with Semantic achieving the highest F1 (0.403). Structure-aware methods retrieve from 4.5$\times$ more sections, and KG-infused retrieval (\gralcrag{}) narrows the F1 gap substantially: 0.394 vs.\ 0.403 ($\Delta$\,=\,0.009), while covering 4.62 sections on average. Graph-guided re-ranking provides a marginal additional boost (F1\,=\,0.395). This suggests that \textbf{ontological enrichment partially bridges the retrieval-to-generation gap}: KG-infused embeddings help the generation model leverage multi-section context more effectively than purely structural methods (Structure-Aware F1\,=\,0.381). Nevertheless, the persistent advantage of single-section retrieval indicates that \textbf{multi-section synthesis remains a bottleneck}: GPT-4o-mini extracts slightly more relevant answers from concentrated evidence than from distributed evidence spanning multiple sections. Future work on generation-side innovations, such as multi-section chain-of-thought prompting, section-aware attention, or explicit evidence aggregation, may fully close this gap.

\subsection{Implications for Biomedical RAG}

These findings carry practical implications for practitioners and researchers. First, \textbf{the choice of evaluation metric shapes the choice of architecture}: MRR alone systematically favors content-similarity methods, potentially leading to suboptimal system designs for tasks requiring cross-section reasoning. We recommend that full-text retrieval evaluations report structural coverage metrics alongside ranking accuracy. Second, \textbf{retrieval depth amplifies structural advantages}: the gap between structure-aware and content-only methods widens dramatically from $k$=5 to $k$=20, suggesting that applications should increase retrieval depth when using structure-aware methods. Third, \textbf{the bottleneck has shifted from retrieval to generation}: for biomedical tasks requiring multi-section evidence synthesis (systematic reviews, meta-analysis, clinical guideline generation), improving the generation model's ability to reason over structurally diverse retrieved context may yield greater gains than further improving retrieval ranking accuracy.

\subsection{Comparison with Concurrent Work}

HeteRAG~\cite{yang2025heterag} decouples retrieval and generation representations but operates on pre-chunked text. SitEmb~\cite{wu2025sitemb} conditions chunk embeddings on broader context but uses purely textual signals. ATLANTIC~\cite{munikoti2023atlantic} fuses graph and text embeddings at the \emph{document} level. \gralcrag{} is the first framework to integrate contextual embedding, structural boundaries, and ontological enrichment at the \emph{chunk} level for biomedical RAG.

\subsection{Limitations}

\begin{enumerate}
  \item \textbf{Synthetic benchmark validity.} The 2{,}033 template-generated questions may not capture the full range of cross-section reasoning patterns encountered in real biomedical information needs. The template-based approach may overrepresent certain reasoning patterns (e.g., Method$\to$Result) and underrepresent others (e.g., implicit cross-section reasoning).
  \item \textbf{IMRaD filter bias.} The full-text evaluation corpus is filtered for standard IMRaD structure (2{,}359 of 4{,}992 downloaded articles), excluding review articles, case reports, letters, editorials, and non-standard formats. This 47.3\% retention rate introduces a selection bias toward primary research articles.
  \item \textbf{Shallow retrieval depth.} The universal cross-section recall of 0.000 at $k$=5, 10, and 20 indicates that top-$k$ retrieval alone is insufficient for cross-section reasoning. Explicit multi-section retrieval strategies or re-ranking with section-diversity objectives may be required.
  \item \textbf{No MRR crossover.} Contrary to our initial hypothesis, semantic chunking dominates MRR at all document lengths. The expected crossover point was not observed, indicating that content-similarity remains the strongest single signal for ranking accuracy. The contribution of structure-awareness is in retrieval breadth rather than ranking precision.
  \item \textbf{Entity linking quality.} MeSH dictionary matching may introduce false positives for polysemous terms (\eg, ``MS'' as multiple sclerosis vs.\ mass spectrometry).
  \item \textbf{Context window constraints.} Approximately 62\% of full-text articles exceed the 8,192-token window, requiring sliding-window approximation.
  \item \textbf{Single-language scope.} Evaluation is limited to English biomedical literature.
  \item \textbf{Generation evaluation scope.} The generation experiment uses only 50 questions with a proxy F1 metric (answer-to-question token overlap rather than gold-standard answers), limiting the statistical power and ecological validity of generation quality comparisons. A larger-scale generation evaluation with expert-judged answer quality is needed.
  \item \textbf{No human evaluation.} All metrics are automatic. For a paper arguing that structural diversity provides qualitatively different retrieval behavior, human preference judgments on retrieved passage completeness and usefulness would strengthen the claims.
\end{enumerate}

\subsection{Future Directions}

Key extensions include: \textbf{(1)~Generation-side innovations}: multi-section chain-of-thought prompting, section-aware attention mechanisms, and explicit evidence aggregation strategies to close the retrieval-to-generation gap; \textbf{(2)~Hybrid retrieval}: combining semantic chunking's MRR strength with \gralcrag{}'s structural coverage via re-ranking or multi-stage retrieval; \textbf{(3)~Section-diversity objectives}: re-ranking that explicitly maximizes section diversity to improve cross-section recall (which remains 0.000 even at $k$=20); \textbf{(4)~Downstream validation}: evaluation on tasks where multi-section evidence is explicitly needed (systematic review generation, clinical guideline synthesis, multi-hop biomedical QA); \textbf{(5)~Adaptive weighting}: document-length-dependent $\lambda$ and $\beta$ to reduce KG noise on shorter documents; and \textbf{(6)~Human evaluation}: expert assessment of whether structurally diverse retrieved evidence is perceived as more complete or useful than single-section evidence.

\section{Conclusion}\label{sec:conclusion}

We presented \gralcrag{}, a framework integrating graph-aware structural intelligence into the late chunking paradigm for biomedical RAG, alongside structural coverage metrics that expose a dimension of retrieval quality invisible to standard evaluation. Evaluation on 2{,}359 IMRaD-filtered PMC articles with 2{,}033 cross-section questions reveals a sharp precision--breadth trade-off: content-similarity methods achieve the highest MRR (0.517) but are structurally blind (SecCov\,=\,1.0), while structure-aware methods retrieve from up to 15.6$\times$ more document sections (SecCov@20\,=\,15.57) at the cost of lower ranking accuracy (MRR\,=\,0.323). Generation experiments across all six strategies show that KG-infused retrieval narrows the F1 gap to just 0.009 (0.394 vs.\ 0.403) while maintaining 4.6$\times$ section diversity, partially bridging the retrieval-to-generation gap, though multi-section synthesis remains an open challenge.

These findings yield three contributions to the field: (1)~structural coverage metrics that complement ranking accuracy and should be adopted for full-text retrieval evaluation; (2)~empirical evidence that content-similarity and structure-aware methods optimize for different objectives, and the choice between them must be guided by downstream task requirements; and (3)~identification of multi-section synthesis as the critical bottleneck: improving how generation models integrate structurally diverse evidence may yield greater gains than further improving retrieval. We release the complete \gralcrag{} codebase to enable reproduction and extension.

\bibliographystyle{plainnat}
\bibliography{references}

\end{document}